\documentclass[10pt,twocolumn,letterpaper]{article}

\usepackage{iccv}
\usepackage{times}
\usepackage{epsfig}
\usepackage{graphicx}
\usepackage{listings} 
\usepackage{amsmath}
\usepackage{amssymb}
\usepackage[caption=false]{subfig} 

\usepackage[pagebackref=true,breaklinks=true,colorlinks,bookmarks=false]{hyperref}

\iccvfinalcopy
\def\iccvPaperID{7805}

\ificcvfinal\fi
\begin{document}

%%%%%%%%% TITLE
\title{Language Models as Zero-shot Visual Semantic Learner}
\author{Yue Jiao\\
University of Southampton\\
Southampton, UK\\
{\tt\small yj5y15@soton.ac.uk}
\and
Jonathon Hare\\
University of Southampton\\
Southampton, UK\\
{\tt\small jsh2@ecs.soton.ac.uk}
\and
Adam Prügel-Bennett\\
University of Southampton\\
Southampton, UK\\
{\tt\small apb@ecs.soton.ac.uk}
}

\maketitle
%\thispagestyle{empty}

%%%%%%%%% ABSTRACT
\begin{abstract}
Visual Semantic Embedding (VSE) models, which map images into a rich semantic embedding space, have been a milestone in object recognition and zero-shot learning. Current approaches to VSE heavily rely on static word embedding techniques. In this work, we propose a Visual Semantic Embedding Probe (VSEP) designed to probe the semantic information of contextualized word embeddings in visual semantic understanding tasks. We show that the knowledge encoded in transformer language models  can be exploited for tasks requiring visual semantic understanding. The VSEP with contextual representations can distinguish word-level object representations in complicated scenes as a compositional zero-shot learner.
We further introduce a zero-shot setting with VSEPs to evaluate a model’s ability to associate a novel word with a novel visual category.
We find that contextual representations in language models outperform static word embeddings, when the compositional chain of object is short. 
We notice that current visual semantic embedding models lack a mutual exclusivity bias which limits their performance. 
\end{abstract}

\section{Introduction}

Visual-semantic embedding models attempt to learn semantic relationships between labels, and explicitly map images into a rich semantic embedding space~\cite{frome2013devise, wang2018zero, liu2020hyperbolic}.
They has been considered key to dealing with novel categories by transferring semantic knowledge obtained from familiar classes. 
However, these models only distill semantic information from static word embeddings~\cite{mikolov2013distributed, pennington-etal-2014-glove}. 
In a static word embedding space, each word has a single vector, regardless of context. 
This constrains all senses of a polysemous word to share the same representation. 

Language models which learn contextualized word representations have revolutionized NLP over the last few years~\cite{peters2018deep, devlin2018bert, 2020t5, brown2020language}.
These models learn highly transferable and task-independent representations. They achieve state-of-the-art performance on various downstream NLP tasks~\cite{brown2020language}. 
They also reveal some interesting behaviors, such as performing remarkably well on open-domain question answering without gradient updates or fine-tuning~\cite{petroni2019language, brown2020language}.

This pretrain-and-finetune scheme has been expanded to the joint domain of vision and language recently,
which demonstrate the potential of transformer-based language modeling, masked language modeling, and contrastive objectives to learn image representations from text~\cite{desai2020virtex,li2020oscar, radford2learning}.  However, there is no evidence that without large-scale modality interaction, visual models can extract semantic information directly
from a pre-trained, frozen language model.
Transferring the knowledge encoded in transformer language models to visual semantic understanding tasks still remain poorly understood.
If visual semantic information is stored in transformer language models, a linear probing model trained on frozen contextualized representations should help distinguish visual concepts.

As language models usually fail when generalization requires systematic compositional skills~\cite{lake2018generalization}, it is important to 
determine whether the probing model still lacks sensitivity to compositionality. On the other hand, capturing semantic relation from a static word embedding space has been shown not to be robust when used to solving zero-shot learning tasks~\cite{hascoet2019zero}.
Rethinking how language shapes the way human learn novel objects is necessary. 
Therefore, we look into  transformer language models and contextual representations to gain deeper insights into how language models help visual semantic understanding tasks by its zero-shot learning capabilities.

We propose the visual semantic embedding probe (VSEP) that aligns representations in two modalities at the word level.
We focus on asking what visual semantic information is encoded in a language model and how well it encodes the compositional structure.
We leverage the image-text pairs in MS-COCO~\cite{lin2014microsoft} to build a visual semantic understanding task. By aligning word representations with object representations in a simple scene, we find that the VSEP with contextual representations can distinguish word-level object representations in a more complicated scene.
The performance is affected by the number of objects in the scene.
We find that normalization is essential for aligning ``anisotropic'' semantic representations.

We also introduce a zero-shot learning task with the VSEP  to  evaluate  a  model’s  ability to associate a novel word with a novel visual category. The contextual representations in the language model outperforms conventional word embeddings, when the number of objects in the scene is low. By analysing the percentage of misclassified samples, we find that current visual semantic embedding models lack a mutual exclusivity bias. The mutual exclusivity bias is what helps children learn the meaning of new words efficiently~\cite{markman1988children}; if a child already knows a label for an object, a new label for that object should be rejected.

These insights help justify the role of contextualized representations in object recognition and zero-shot learning.

\section{Background}
In this section, we provide background on learning semantic representations in 
language and visual modalities.
Then we introduce the open generalization problems in multi-modal embedding models.

\begin{figure*}
\begin{center}
\includegraphics[width=\linewidth]{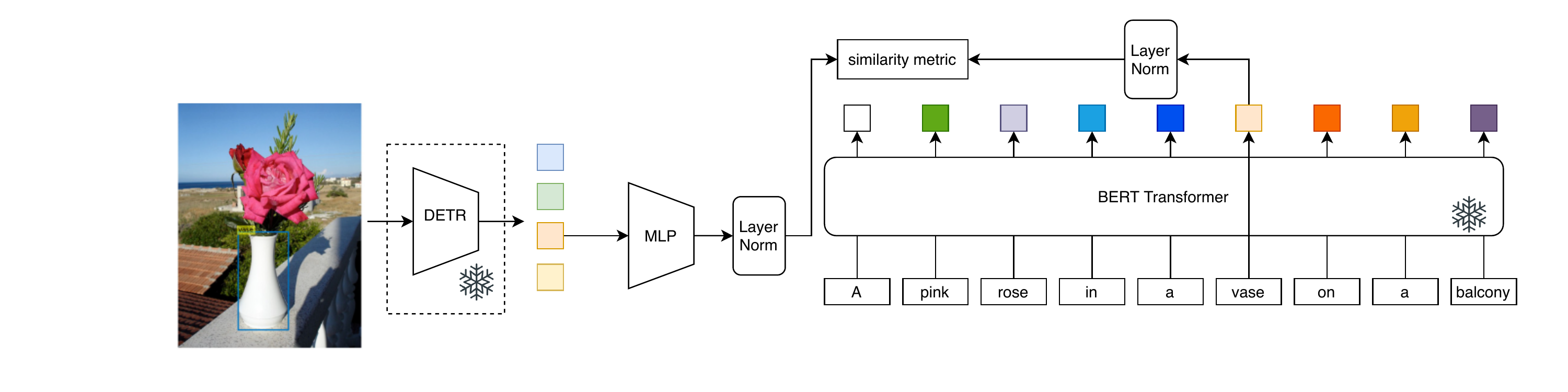}
\end{center}
  \caption{Illustration of the Visual Semantic Embedding Probe (VSEP). Word-centric semantic representations from two frozen embedding systems are aligned after normalized by Layer Normalization. Our aim is to explore to what extent two independent embedding systems in different modalities can interact.  It is the key technique to transfer semantic knowledge from a language model to a visual model.}
\label{fig:pipeline}
\end{figure*}

\subsection{Word Representations}
\begin{itemize}
\item \textbf{Distributed Word Representations.}
\textbf{GloVe}~\cite{pennington-etal-2014-glove} is a fast and efficient algorithm for generating distributed word representations, often considered points in a semantic space.
Given input word pairs $\{(w_{i}, c_{j})\}$ extracted from a large text corpus, 
target word $w_{i}$ ranges over the corpus and context word $c_{j}$ ranges over a sliding window of size $l$ which is symmetric about $w_{i}$.
We denote the number of observed word pairs as $D$. 
For each observed word pair, $k$ random negative samples are generated from unigram distributions.
We use $\#(w_{i}, c_{j})$ to denote the number of times the pair $(w_{i}, c_{j})$ appears.
$\#(w_{i})$ and $\#(c_{j})$ are the number of times $w_{i}$ and $c_{j}$ occurred respectively.
For embedding dimension $d$ and dictionary size $n$, GloVe comprises the product of two weight matrices $\mathbf{W}, \mathbf{C} \in \mathbb{R}^{d\times n}$ subject to the logistic sigmoid function. 
The loss function of GloVe is given by:
\begin{equation}
 \begin{split}
L_{GloVe} =- \sum_{i=1}^{n}\sum_{j=1}^{n}&\#(w_{i}, c_{j})\log\sigma (\mathbf{w_{i}}^\top \mathbf{c_{j}})\\
 &+\frac{k}{D}\#(w_{i})\#(c_{j})\log\sigma (-\mathbf{w_{i}}^\top \mathbf{c_{j}})
 \end{split}
 \end{equation} 
where $\mathbf{w_{i}}$ and $\mathbf{c_{j}}$ are columns of $\mathbf{W}, \mathbf{C}$. This function is minimised when,
\begin{equation}
\mathbf{w_{i}}^\top \mathbf{c_{j}} = \log(p(w_{i}, c_{j})) -b_{i} -b_{j}+\log(Z)
\end{equation}
with $Z$ representing a normalising constant.
Although GloVe can learn statistical semantic relationships flexibly, such as similarities and analogies\cite{allen2019vec}, it makes some assumptions about language that do not fit with reality. 
Most importantly, it can not capture different meanings in different contexts \cite{smith2019contextual}.

\item \textbf{Contextual Word Representations.}
\textbf{BERT} is a bidirectional language model which can learn contextual representations of words.
Formally, given an input sequence $\{w_{1},w_{2}, ... ,w_{N}\}$, we want to estimate $p(w_{i})$ using the left and right context of $w_{i}$.
To this end, BERT employs a deep Transformer \cite{vaswani2017attention} encoder to learn to fill the word at masked positions and to predict next sentences, trained on the concatenation of the Toronto Books Corpus \cite{zhu2015aligning} and English Wikipedia.

It has been pointed out that BERT produces strong representations for syntactic phenomena \cite{tenney2019you}, and contains relational knowledge competitive with traditional NLP methods that have some access to oracle knowledge~\cite{petroni2019language}.
Most interestingly, BERT also yields representations that are useful in retrieving semantically aligned image patches \cite{ilharco2020probing}.

\item \textbf{Anisotropic Semantic Space.}
The ``Anisotropic'' property means learned representations are not uniformly distributed with respect to direction. This property makes it hard to use word representations directly through simple similarity metrics.
Both distributed word embedding models and language models have been found to induce non-smooth anisotropic semantic spaces, which harms their performance of semantic similarity \cite{mu2017all, gao2019representation, ethayarajh2019contextual}. 
\end{itemize}

\subsection{Visual Object Representations in Context}
Object detection can be seen as a set prediction problem~\cite{carion2020end}.
A pre-trained object detection model can predict a set of object representations that contain rich contextual information.
In this work, we employ a pre-trained object detection module, DETR \cite{carion2020end}, to extract contextual visual features.
Using self and encoder-decoder attention over object queries embeddings, the model globally reasons about all objects together using pair-wise relations.  It then predicts visual object representations before the final feed-forward networks.

\subsection{Visual Semantic Embedding Models}
A visual semantic embedding model is built to align semantics in different modalities. 
Formally, given a set of image-word pairs, $D=\{(x,y) \mid x \in X, y \in Y\}$, a semantic alignment learner finds structural correspondence between the two embedding systems.

The \textbf{Deep VIsual Semantic Embedding model, DeVISE}, proposed by Frome \etal~\cite{frome2013devise}, is a milestone in visual semantic embedding models. It maps visual features to the word embedding space by a combination of dot-product similarity and hinge rank loss.
 For a data instance in $D$, we have the loss:
 \begin{equation}
 \begin{split}
 &L_{DeVISE}(x, y) =  \sum_{j\neq y}\max\bigl[0, \\
 &\hspace{1cm} margin - w(y)\cdot t(f(x)) +w(j)\cdot t(f(x))\bigr]
 \end{split}
 \end{equation}
where $t$ is a trainable transformation neural network, $f$ and $w$ are visual and language feature extractors. 

This loss aims to make projected image features close to corresponding word representations while remaining distant from negative word embeddings.
The DeVISE model is a natural way to project the information in the visual domain to the distributed word space. However, it also can not capture different meanings of words in different contexts.

\subsection{Generalization in Visual Semantic Understanding}

\begin{itemize}
\item \textbf{Compositional Generalization.}
Compositionality is one of the features shared by many human designed representation systems. 
It is the capacity to represent complex concepts by combining simple parts \cite{fodor2002compositionality}.
In language understanding, for example, if a person knows the meaning and usage of words such as ``twice'', once she learns a new action called ``dax'', she can immediately understand or produce instructions such as ``dax twice'' \cite{lake2018generalization}.
In this work, we explore the compositional skills of language models in visual semantic understanding through a series of novel classification tasks.

\item \textbf{Zero-shot Generalization.}
Zero-Shot Learning (ZSL) aims to recognize unseen classes. ZSL is an important task for demonstrating how a machine learning model understands high-level semantic information and transfers knowledge from seen to unseen classes.

Recent ZSL models~\cite{frome2013devise, xian2017zero, schonfeld2019generalized, hascoet2019zero} encode semantic information from distributed word representations, such as Word2vec \cite{mikolov2013distributed}, GloVe \cite{pennington-etal-2014-glove} and Poincaré Embeddings \cite{nickel2017poincare}. 
Our investigation seeks to explore to what extent pre-trained language models store visual semantic knowledge.
\end{itemize}

\section{Visual Semantic Embedding Probe}
We introduce the Visual Semantic Embedding Probe (VSEP) to test the visual semantic knowledge in language models.
Our approach (see Figure~\ref{fig:pipeline}) uses a shallow neural model that maps frozen visual region representations to word-level language representations. 
To solve the problem caused by anisotropic word representations, we add an additional layer normalization
before the similarity metrics.

Given a batch of $N$ visual region and word representation pairs, our models learns a
visual semantic embedding space by training a shallow neural network to maximize the cosine similarity between the mapped visual region representations and the word representations of the $N$ real pairs in the batch while minimizing the cosine similarity of the embeddings of the $N(N-1)$ incorrect pairings.
This training technique has been popularized for multi-class N-pair loss and contrastive representation learning~\cite{NIPS2016_6b180037, oord2018representation, radford2learning}. Following what was done in \cite{radford2learning}, we also apply a learnable temperature $T$, which is directly optimized during training to avoid turning as a hyper-parameter. The pseudocode of training procedure is in Figure~\ref{fig:code}.

Once trained, the VSEP can map arbitrary visual region representations to a language semantic space.  It can then be used to evaluate 
to what extend learned linguistic knowledge in distributed word models and contextual language models helps generalization through visual semantic embedding.

Although visual semantic embedding is not a novel framework, the VSEP extends the original idea of DeVISE to align visual regions with contextual linguistic information, which provide a method to understand the role of pre-trained language model in visual semantic understandings.
Unlike CLIP~\cite{radford2learning}, VSEP is not a multimodal pre-training technique and has very different aims. In particular we wish to use VSEP to explore to what extent two independent embedding systems in different modalities can interact.
It is a technique to provide supplementary semantic knowledge transferred from
language domain to a visual understanding system. This differs from capturing high-quality multimodal representations from raw data.  Training a VSEP does not demand large amounts of raw image and text pairs. 
We focus on the semantics of the object-centric connections between the language embeddings and the vision embeddings.

Closely related to our work is that done by Ilharco \etal~\cite{ilharco2020probing}. Compared with that work, VSEP solves the problem caused by anisotropic semantic representations and provides a set of fair zero-shot learning settings which consider the compositional generalization behavior and mutual exclusivity bias.

\begin{figure}[tbp]
\begin{lstlisting}
# I[n, v] 
# - minibatch of representations of image regions
# T[n, l] 
# - minibatch of word-level representations
# MLP - learned projection
# t - learned temperature parameter
I_e = np.linalg.norm(MLP(I), axis=1)
T_e = np.linalg.norm(T, axis=1)
# scaled pairwise cosine similarities [n, n]
logits = np.dot(I_e, T_e.T) * np.exp(t)
# symmetric loss function
labels = np.arange(n)
loss_i = cross_entropy_loss(logits, labels, axis=0)
loss_t = cross_entropy_loss(logits, labels, axis=1)
loss = (loss_i + loss_t)/2
\end{lstlisting}
\caption{Numpy-like pseudo-code for the VSEP training procedure. Unlike the CLIP, which uses a input pair of consisting of an image feature and embedded semantic information, the input of VSEP consists of pairs of object-centric feature representations and their associated semantic embedding.}
\label{fig:code}  
\end{figure}

\section{Experiment Settings}
In this section we discuss the experimental details of our approach, including the data, models and tasks.

\subsection{Creating Visual Region Pairs}
MS-COCO~\cite{lin2014microsoft}, Visual Genome~\cite{krishna2017visual} and Flickr30k~\cite{fukui2016multimodal}
provide high quality labeled object-centric semantic information.  They are mainly used in object detection~\cite{ren2016faster, carion2020end}, image-text embedding~\cite{faghri2017vse++, wu2019unified} and image captions~\cite{vinyals2015show, lu2018neural}.
Whilst in Visual Genome and Flickr30k, object categories are defined as the set of unique phrases, only MS-COCO has clear object categories and complete sentences describing whole images. We therefore use the MS-COCO dataset in our experiments.

We use the official implementation of the DETR~\cite{carion2020end} detector to generate object representation of 80 categories.
To avoid having multiple representations for one class, we choose the representation with the highest probability per object category in each image.

Following the setting in Neural Baby Talk~\cite{lu2018neural}, we choose 413 fine-grained classes of the original 80 categories.
For each image, we choose one sentence which contains the most fine-grained classes.
We collect the GloVe embedding and BERT embedding for the fine-grained classes in each sentence.

The data collection process described above is a fairer test than the process described in~\cite{ilharco2020probing}. 
That process used 1600 object categories defined by Anderson \etal~\cite{Anderson2017up-down}.  
However, we notice that some categories are not in MS-COCO captions and as such, we do not choose the detector trained on Visual Genome.

\begin{figure}[htb]
 \centering
 \subfloat['a pink rose in a \textcolor{red}{vase} on a\\ balcony',]{\includegraphics[height=3cm]{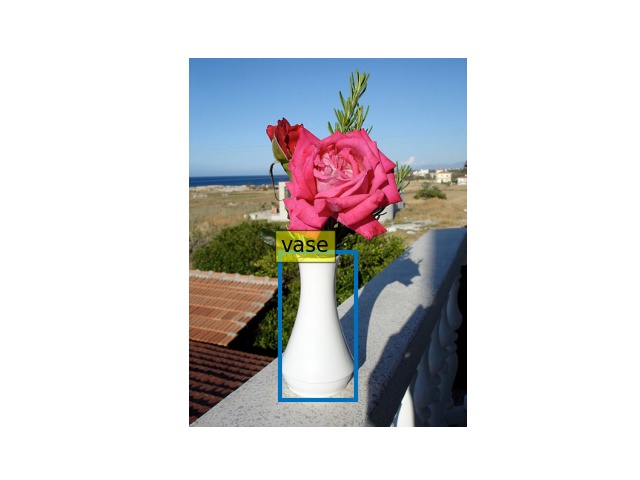} 
 \label{fig: vase}}%
 \subfloat['a tall \textcolor{blue}{clock} tower with a circular clock\\ under a blue sky']{\includegraphics[height=3cm]{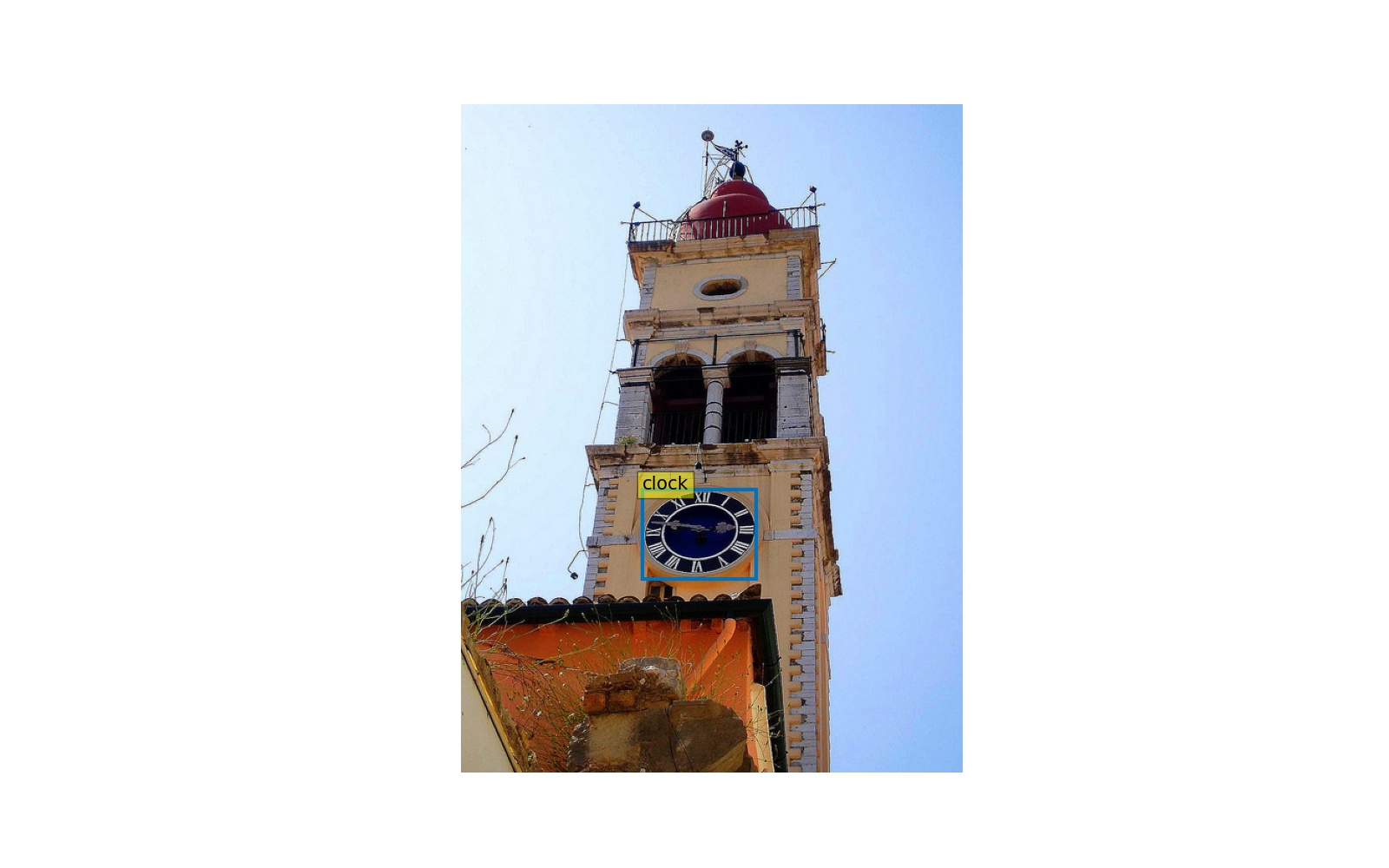}
 \label{fig:clock}}\\
 \subfloat['a \textcolor{red}{vase} with flowers, a \textcolor{blue}{clock} and\\ paintings inside of a building']{\includegraphics[height=3cm]{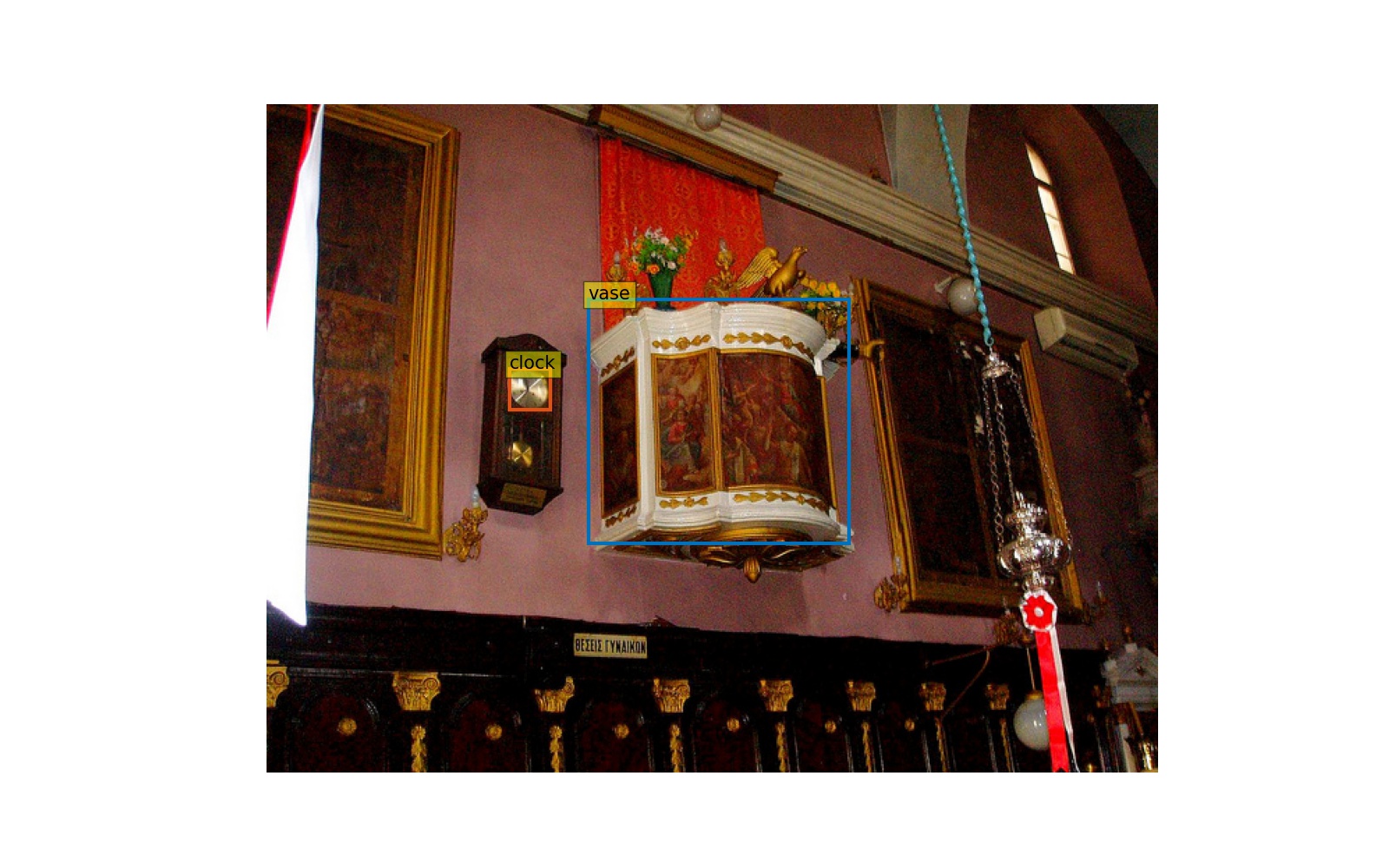}  
 \label{fig:vase_clock}}%
  \caption{The top images contain a vase and a clock respectively. The bottom image and a caption are testing pairs. The basic visual semantic understanding task is to predict the right visual regions given a set of word representations.}
 \label{fig: task}%
 \end{figure}

\subsection{The Basic Object-Label Alignment Task}
Since the fine-grained classes in captions are chosen as semantic concepts, we should avoid structural flaws in evaluation~\cite{hascoet2019zero}.
When classes that are hypernyms or hyponyms of other classes within the WordNet~\cite{miller1995wordnet} hierarchy are used in a multi-way classification task to evaluate the performance of a visual semantic embedding model it is likely to give a two optimistic view of its generalization performance.
For example, representations of ``dog'' and ``puppy'' should not appear in one evaluation scene.
Therefore, we design a basic task to demonstrate how visual semantics are aligned with a language model.

We firstly choose images which contain only one object from the training and the validation sets of MS-COCO.
Visual and language representation pairs in these images are used to train the VSEP. 
There are 51,637 images in the training set. The testing set is partitioned into three independent subsets of increasing difficulty: (i) 53,378 images which have two objects, (ii) 9941 images which have three objects and (iii) which has 901 images with four objects. The subsets were created by counting the number of objects using the pre-trained DETR network.

With the VSEP, we calculate the similarity matrix of the projected visual representations and the word representations in each testing image, and predict their relations.
We calculate the accuracy defined as the percentage of the representation pairs in which visual representations are correctly matched with their language representations.
The accuracy of relation prediction is chosen as the metric to avoid the problem of a multi-way classification task.
Figure~\ref{fig: task} shows an episode of the basic task. In this example, we make the VSEP learn the alignment between image regions and word representations of ''vase'' and ''clock''. Given a novel scene which contains both ''vase'' and ''clock'', the model should distinguish the provided visual representations.
Note that the testing sentences are composed of objects in training data, this task is a natural way to detect systematic compositional skills of the VSEP.

Inspired by experiments in CLIP, we also test word embeddings from a simple caption template: \verb|``A photo of a {label}''|. Note that these captions do not contain any context. We believe that this comparison can help us to investigate whether contextual information is essential for VSEPs.

Note that this basic task is natural to evaluate a model’s ability to associate a word with a visual category  and immediately use that word in an compositional zero-shot way.
\subsection{The Zero-shot Object-Label Alignment Task}
We test our VSEP in a zero-shot learning task to demonstrate whether a language model can be used as a visual semantic knowledge base.
We replicated the experimental design on training visual-language representation pairs which exclude at least one of eight objects in COCO. The excluded objects are fine-grained classes of ‘bottle’, ``bus'', ``couch'', ``microwave'', ``pizza'', ``racket'', ``suitcase'' and ``zebra''.
This split is usually applied in novel object captioning tasks.
For testing, we collect the images which include these objects in three testing sets respectively.

Note that in~\cite{ilharco2020probing}, 1,600 classes are randomly split into a seen and unseen set. However, there are a lot of
synonyms in the 1,600 categories. Our split guarantees that objects in the testing dataset are not leaked to the VSEP.

\subsection{Image Patch Retrieval}
We compute instance recall in retrieving image patches given object representations of the same category in different text.
From the images which contain two objects in MS-COCO, we collect 100 images for each category (80 categories in total).
The instance recall (IR@k) is the average percentage of pairs $(v,l)$ in each category where the instance $v$ is in the top $k$ visual representations retrieved from a language representation $l$.
We believe this task can demonstrate to what extent a VSEP can understand context in language precisely.

\subsection{Training Details}
For the visual object encoder, we use the pre-trained DETR network with a ResNet-50, obtaining 42 AP on MS-COCO \footnote{https://github.com/facebookresearch/detr}.
For the static word embeddings, we choose 300 dimensional GloVe\footnote{https://nlp.stanford.edu/projects/glove/} with 6 billion tokens as the pre-trained word vectors. For each fine-grained class, we average all the Glove vectors of its words.
For the contextual language model, we choose the BERT base model trained by Hugging Face\footnote{https://huggingface.co/bert-base-uncased}.

We build the VSEP with PyTorch Lightning~\cite{falcon2019pytorch}.
The VSEP has a two-layer multi-layer perceptron (MLP) with 512 hidden neurons. All the pre-trained models are frozen. Adam with learning rate $1\times10^{-3}$ is used in all the models to optimize the loss function. All the models are trained for 200 epochs with mini-batches of size 512.

\section{Results}
\subsection{The Effect of Anisotropy}
We firstly demonstrate the phenomenon that anisotropic language embedding space induces poor performance in visual semantic similarity comparison.

\begin{figure}[htb]
 \centering
 \subfloat[GloVe]{\includegraphics[width=0.2\textwidth]{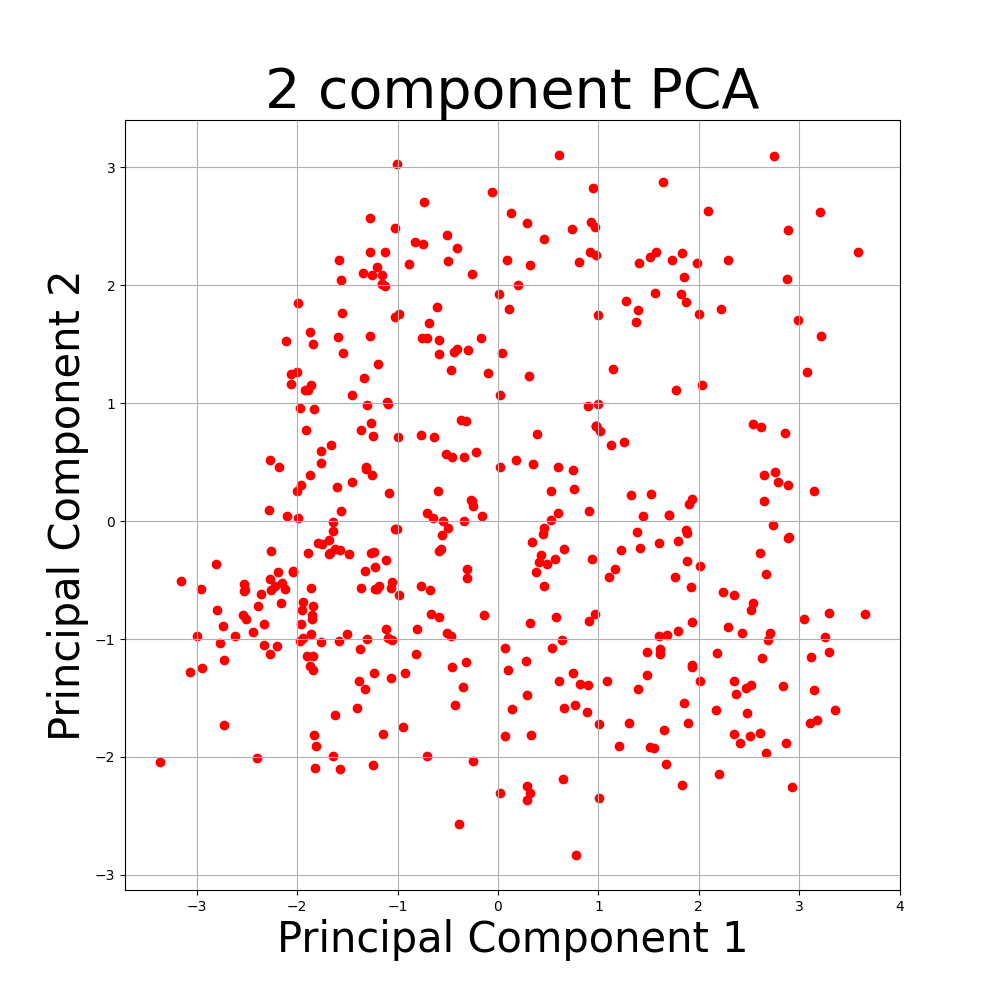} 
 \label{fig:a}}%
 \subfloat[BERT]{\includegraphics[width=0.2\textwidth]{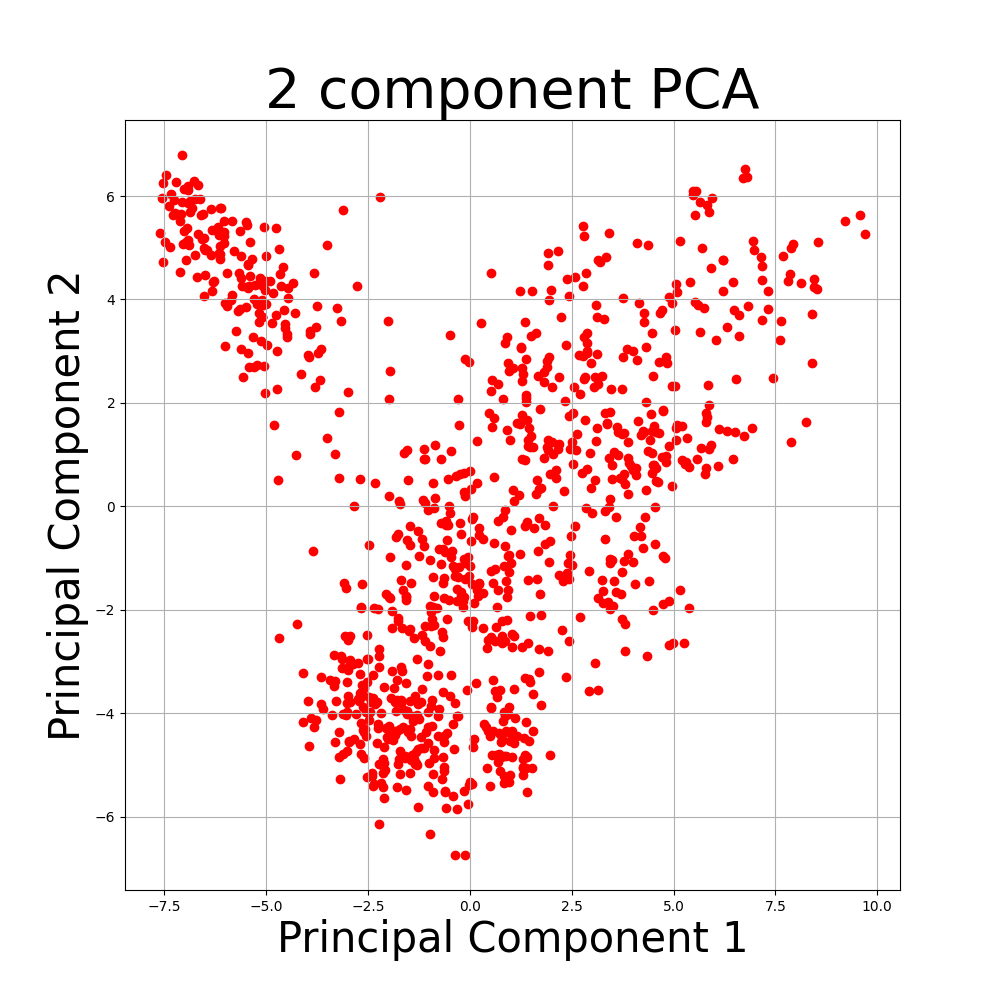}  
 \label{fig:b}}\\
  \caption{Visualization of word embeddings in GloVe and BERT. The GloVe embeddings have non-zero mean. The BERT embeddings are far from isotropic.}
 \label{fig: embedding}%
 \end{figure}

\begin{table*}
\begin{center}
\begin{tabular}{l|l|l|l|l}
\hline
Embeddings & GloVe & GloVe with LN & BERT & BERT with LN \\
\hline\hline
2 Objects & 49.93\%& 99.68\% & 52.29\%& 98.72\%        \\

3 Objects & 32.87\%& 99.15\% & 35.95\%& 97.31\%     \\

4 Objects & 24.50\%& 99.36\% & 27.80\%& 97.14\%       \\
\hline
\end{tabular}
\end{center}
\caption{Layer normalization renders stronger word representations in visual semantic embedding probes: the table show that the matching accuracies are 
significantly improved with a layer normalization before similarity calculation.}
\label{table:1}
\end{table*}

In Figure~\ref{fig: embedding}, we show the GloVe embeddings of the 413 fine-grained classes and 1,000 sampled BERT embeddings corresponding to all the classes in a 2D plane spanned by the largest two principle components.
Word representations in BERT are obviously anisotropic and 
they all occupy a narrow part in the vector space rather than being uniform in all directions.

Meanwhile, the energy of most word vectors is contained in a very low dimensional subspace.
Note that these embedding spaces are of a small part of vocabularies in GloVe and BERT.
The space of GloVe embeddings has many holes, while the space of BERT embeddings is dense but narrow.

We use layer normalization (LN) \cite{ba2016layer} to rescale word embeddings before we calculate semantic similarity.
The results are shown in Table~\ref{table:1}, both GloVe and BERT embeddings work well with layer normalization.
Their scores are significantly better than the models without layer normalization.
We notice that the VSEP without layer normalization is close to the expected performance of random guessing ($50.00\%$ for 2 Objects, $33.33\%$ for 3 Objects and $25.00\%$ for 4 Objects). 
The results demonstrate that layer normalization can mitigate the problem of anisotropic embeddings in visual semantic alignment.
In the following experiments, we use layer normalization in all VSEPs.

\subsection{Knowledge in Language Models}
We compare the performance of static representations with contextual representations.
Besides GloVe embeddings, we also use random unit-norm embedding vectors as embeddings for the 413 fine-grained classes.
The results are shown in Table~\ref{table:2}. All three kinds of embeddings work well with the VSEP.
The static embeddings perform slightly better. 
We observe that contextual representations do not show the benefit in the task when all the objects are well defined before visual semantic alignment.
However, contextual representations still show their ability to deal with unseen compositional scenes and sentences. 

We also compare the performance when we use a label template \verb|``A photo of a {label}''| for testing semantic information.
For GloVe embeddings, we average all the vectors in the template. For BERT embeddings, the template is considered as the input.
The results in Table~\ref{table:3} show that the new template influences the effect of VSEP aligning visual semantics.
Compared with BERT, the static word embeddings show a more robust performance.
This demonstrates language models still rely on contextual information to inference similarities, even though input information is from another modality.

\begin{table}
\begin{center}
\begin{tabular}{l|l|l|l}
\hline
Embeddings &Random Vectors & GloVe & BERT \\
\hline\hline
2 Objects & 99.41\%& 99.68\% & 98.72\%        \\

3 Objects & 99.13\%& 99.15\% & 97.31\%     \\

4 Objects & 99.03\%& 99.36\% & 97.14\%       \\
\hline
\end{tabular}
\end{center}
\caption{The performance of VSEPs with static and contextual word embeddings. The results show that static embeddings perform slightly better when all the objects are well defined before visual semantic alignment. The language model is also capturing useful information.}
\label{table:2}
\end{table}

\begin{table}
\begin{center}
\begin{tabular}{l|l|l}
\hline
Embeddings & GloVe & BERT \\
\hline\hline
2 Objects & 0.29\% & 6.86\%        \\
\hline
3 Objects & 0.51\% &2.27\%     \\
\hline
4 Objects & 0.11\% & 0.36\%       \\
\hline
\end{tabular}
\end{center}
\caption{The performance drop with a novel caption template:``A photo of a {label}'' as testing label. Note  that  these  captions do not contain any context. Compared with BERT, the static word embeddings show a more robust performance. These demonstrate that contextual  information  captured  by  BERT  does influence visual understanding.}
\label{table:3}
\end{table}

Our experiments show that visual semantic embedding probes trained with word-level representations in language models are still sensitive to context.
They can work well on tasks which require compositional generalization and have a significant performance decrease with unseen and non-contextual text.
In contrast, the static word representations are more robust.
All these phenomena demonstrate that contextual information as language knowledge in language models still influences a visual semantic embedding model.
However, contextual representations do not show any benefit in a complete information game, when all the label information is provided.

\subsection{Zero-shot Learning with VSEPs}
Next we test the VSEPs with contextual representations in a zero-shot learning task to explore whether other kind of knowledge in a language model is learned by a VSEP.
As we introduced in last section, objects in fine-grained classes of ‘bottle’, ``bus'', ``couch'', ``microwave'', ``pizza'', ``racket'', ``suitcase'' and ``zebra'' do not appear in the training visual and language representation pairs.
For testing, we collect the images which include these objects (5675 images which have two objects, 2517 images which have three objects, 342 images which have four objects).

We firstly evaluate VSEP zero-shot learning performance by calculating region prediction accuracy.
Note that in one image, we extract a set of visual and language representation pairs which contains both seen objects and unseen objects.
The task is to label these image regions from the provided language representations.
The results in Table~\ref{table:4} show that random static vectors do not provide enough information to make the VSEP label novel visual regions.
Using representations from GloVe and BERT, VSEPs can discern image patches from a scene which contains novel objects and novel instructions.
We observe that contextual language representations perform better in the scenes which contains two kinds of objects.
With a longer compositional chain, the static word representations
perform better, showing that the quality of contextual word representations are easily affected  by the length of captions.

Next we only focus on the unseen objects. 
Table~\ref{table:5} shows the labeling error rate for novel regions.
We observe that the VSEP with BERT representations produces less labeling errors. 
The more objects are in one scene, the more mistakes are made by the VSEP.
We believe this phenomenon is related to the mutual exclusivity (ME) bias \cite{au1990principle}.
When children endeavour to learn a new word, they rely on inductive biases
to narrow the space of possible meanings: they prefer to predict that the novel word refers to the novel object.
However, deep learning algorithms lack this bias \cite{gandhi2019mutual}.
To demonstrate this assumption, we calculate the percentage of novel object in the wrongly labeled vision regions.
For the VSEP with GloVe embeddings, it is $98.54\%$, while the number is $89.46\%$ for BERT embeddings.
The VSEPs prefer to predict the novel object to a familiar word, showing a probe bridging two pre-trained embedding systems does not naturally reason by mutual exclusivity.

\begin{table}
\begin{center}
\begin{tabular}{l|l|l|l}
\hline
Embeddings &Random Vectors & GloVe & BERT \\
\hline\hline
2 Objects & 65.31\%& 84.96\% & 86.95\%        \\

3 Objects & 68.94\%& 85.15\% & 84.84\%     \\

4 Objects & 74.63\%& 86.62\% & 83.19\%       \\
\hline
\end{tabular}
\end{center}
\caption{The accuracy of VSEPs in zero-shot learning tasks. The learned embeddings work considerably better than random embeddings, showing that the language models are capturing useful information. The contextual information captured by BERT does help improve performance when there are 2 objects, but performs worse with more than 2 objects.}
\label{table:4}
\end{table}

Summarising the analysis of the zero-shot learning tasks,
we find that the VSEPs with contextual word representations can be used to do inference for unseen objects, which 
means knowledge in language models can boost visual tasks.
However, representation from language models still have their limitations.
Although, VSEPs with them perform better in some scenes, the models still struggle with images which have a long compositional description.
Meanwhile, all the VSEPs with different word embedding models can not learn the mutually exclusivity bias.

\begin{table}
\begin{center}
\begin{tabular}{l|l|l|l}
\hline
Embeddings &Random Vectors & GloVe & BERT \\
\hline\hline
2 Objects & 31.42\%& 70.42\% & 76.17\%        \\

3 Objects & 10.55\%& 57.39\% & 61.10\%     \\

4 Objects & 5.83\%& 51.39\% & 48.06\%       \\
\hline
\end{tabular}
\end{center}
\caption{The percentage of correctly labeled  unseen object regions. 
The learned embeddings work better than random embeddings. The contextual  information helps produces less labeling errors.  We notice there is a significant performance drop with the number of objects in one scene increasing.}
\label{table:5}
\end{table}

\subsection{Analysis of Image Patch Retrieval}
We further investigate the influence of context by analyzing the performance of image patch retrieval with VSEPs.
Previous work \cite{ilharco2020probing} only evaluated the performance of visual instance retrieval with different seen/unseen category splits.
We design a novel image patch retrieval task.
In our approach, all the testing categories are seen by the VSEP.
We prepare 80 independent testing sets for 80 classes in MS-COCO.
Each set is composed by 100 images which are randomly chosen from the 2-Object testing dataset.
The task is to recall image patches given word representations extracted from captions which contains the same object in different scenes.
We evaluate this instance retrieval task for 5 times.
Although this task is harder, it is more able to demonstrate to what extent a VSEP obtain contextual information from a pre-trained language model. 

\begin{figure*}
\begin{center}
\includegraphics[width=\linewidth]{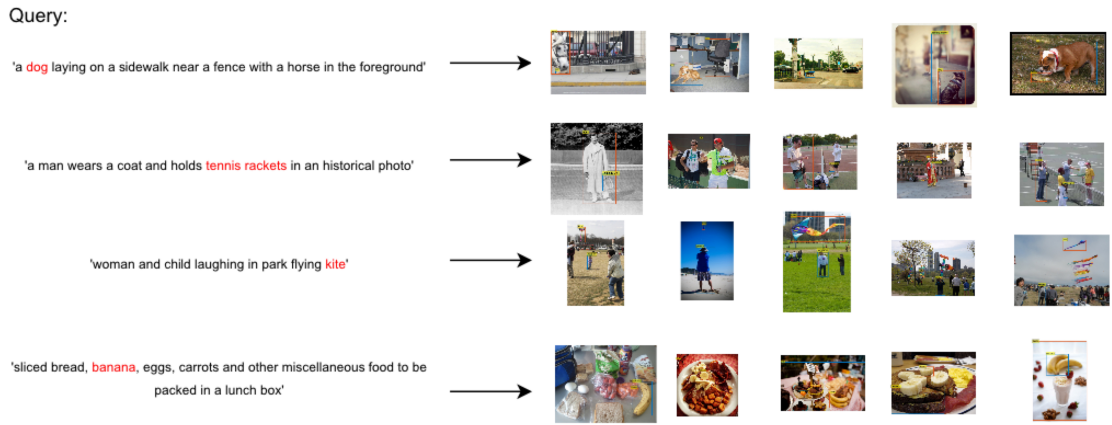}
\end{center}
  \caption{Qualitative examples of top5 images retrieved from contextual representations of objects in captions. All the retrieved images belong to a same category. This task is designed to evaluate how contextual information influences instance retrieval.}
\label{fig:query}
\end{figure*}

As is shown in Table~\ref{table:6}, representations from BERT has a higher performance compared with random picking.
Figure~\ref{fig:query} shows qualitative examples of top5 retrieved images given a contextual word representation (representation of the red word) as query. We observe that most retrieved images have not only the query object, but also the object appears in the whole caption.
For instance, the retrieved images in the fourth row both have a carrot.

It is clear that contextual word representations in language models can provide more precise semantic information.
For visual understanding tasks involving both local and global semantic information in text, static embedding models can not work, using language models is the unique choice.

\begin{table}
\begin{center}
\begin{tabular}{l|l|l}
\hline
 & IR@1 & IR@5 \\
\hline\hline
Random & 1.00\% & 5.00\%        \\

BERT& $4.85\% \pm 0.18\%$ & $17.48\% \pm 0.17\%$           \\
\hline
\end{tabular}
\end{center}
\caption{The average instance recall rate for image patches in a same category. The contextual information captured by BERT does help improve performance compared with random picking. The static word representations can not work on this task because all the visual regions share one word vector.}
\label{table:6}
\end{table}

\section{Conclusion}
We present a simple visual semantic embedding probe designed to probe the semantic information of contextualized word embeddings in a series of visual semantic understanding tasks.

Based on our analysis, we find evidence suggesting the following trends: 
\begin{itemize}
\item Firstly,language models are naturally zero-shot visual semantic learners. They can associate a word with a visual category and immediately use that word in an compositional zero-shot way.

\item Secondly, the performance of contextualized word embeddings is affected by the number of objects in the scene, which indicates that visual semantic embedding models and language models also struggle to generalize by systematic composition.

\item Thirdly, semantic representation alignment requires normalization, as evidenced by the need for a layer-norm to get good results.

\item Fourthly, visual semantic models with contextualized word embeddings and static word embeddings both lack a mechanism to capture a mutual exclusivity bias.
\item Finally, contextualized embeddings can be used to retrieve fine-grained visual content.
\end{itemize}
All of our the results show that knowledge in language models can be exploited in the task of visual semantic understanding to some extent. On the other hand, contextual information in current language models is still not strong enough for zero-shot predictions, and models still prefer to map novel inputs to familiar outputs. 

\section{Further Discussion}
There are still many limitations to our work. 
The main limitation is that captions in MS-COCO do not contain rich semantic information to represent a large amount of objects.
We can just use 413 fine-grained class names in 80 categories.
For multimodal problems, a sufficiently large dataset is the game changer~\cite{radford2learning}.
A large dataset with high-quality annotated captions is required to explore the role of language models in visual tasks.
Current captions in MS-COCO are too simple to describe the complex scenes.

On the other hand, although we design a series of reasonable tasks to compare the static and contextual word representations in a visual semantic embedding system, all the tasks are implicit. We make the models match seen or unseen image regions with words in scene composed of multiple objects.
We also make the models do instance retrieval in the same categories. 
These tasks are all based on the assumption that knowledge in language models should influence the visual understanding tasks.
We can see there is an improvement using contextual word information, but the process is still like a black box. 
There is still a long way to explore what kind of knowledge is stored in a pre-trained language model.

{\small
\bibliographystyle{ieee}
\bibliography{vsep}
}
\end{document}